\pdfoutput=1

\documentclass[11pt]{article}

\usepackage{EMNLP2022}

\usepackage{times}
\usepackage{latexsym}
\usepackage{enumitem,kantlipsum}
\usepackage{graphicx}
\usepackage{caption}
\usepackage{amsmath}
\usepackage{amssymb}
\usepackage{pgfplots}
\pgfplotsset{compat=newest}
\usepackage{comment}
\usepackage{booktabs}
\usepackage{hyperref}
\usepackage[overload]{empheq}
\usepackage{makecell}
\usepackage[ruled,vlined]{algorithm2e}
\usepackage{comment}
\usepackage{arydshln}
\usepackage{array,multirow}
\usepackage{url}

\newcommand{\STAB}[1]{\begin{tabular}{@{}c@{}}#1\end{tabular}}

\usepackage[T1]{fontenc}

\usepackage[utf8]{inputenc}

\usepackage{microtype}

\title{Gaining Insights into Unrecognized User Utterances \\ in Task-Oriented Dialog Systems}

\author{
	Ella Rabinovich\hspace{1.2cm}
	Matan Vetzler\hspace{1.2cm}
	David Boaz\hspace{1.2cm}
	Vineet Kumar
	\vspace{0.1cm} \\
	\textbf{Gaurav Pandey}\hspace{4cm} 
	\textbf{Ateret Anaby-Tavor} 
	\vspace{0.15cm} \\
	IBM Research \\
	\texttt{\{ella.rabinovich1, matan.vetzler\}@ibm.com} \\
	\texttt{\{vineeku6, gpandey1\}@in.ibm.com} \\
	\texttt{\{davidbo, atereta\}@il.ibm.com}
}

\begin{document}
\maketitle
\begin{abstract}
The rapidly growing market demand for automatic dialogue agents capable of goal-oriented behavior has caused many tech-industry leaders to invest considerable efforts into task-oriented dialog systems. The success of these systems is highly dependent on the accuracy of their intent identification -- the process of deducing the goal or meaning of the user's request and mapping it to one of the known intents for further processing. Gaining insights into unrecognized utterances -- user requests the systems fail to attribute to a known intent -- is therefore a key process in continuous improvement of goal-oriented dialog systems.

We present an end-to-end pipeline for processing unrecognized user utterances, deployed in a real-world, commercial task-oriented dialog system, including a specifically-tailored clustering algorithm, a novel approach to cluster representative extraction, and cluster naming.  We evaluated the proposed components, demonstrating their benefits in the analysis of unrecognized user requests.

\end{abstract}

\section{Introduction}

The development of task-oriented dialog systems has gained much attention in both the academic and industrial communities over the past decade. Compared with open-domain dialog systems aimed at maximizing user engagement \citep{huang2020challenges}, task-oriented (also referred to as goal-oriented) dialog systems help customers accomplish a task in one or multiple domains \citep{chen2017survey}. A typical pipeline system architecture is divided into several components, including a natural language understanding (NLU) module, which is responsible for classifying the first user request into potential \textit{intents}, performing a decisive step that is required to drive the subsequent conversation with the virtual assistant in the right direction.

\begin{figure}
\centering
\resizebox{0.95\columnwidth}{!}{
\includegraphics{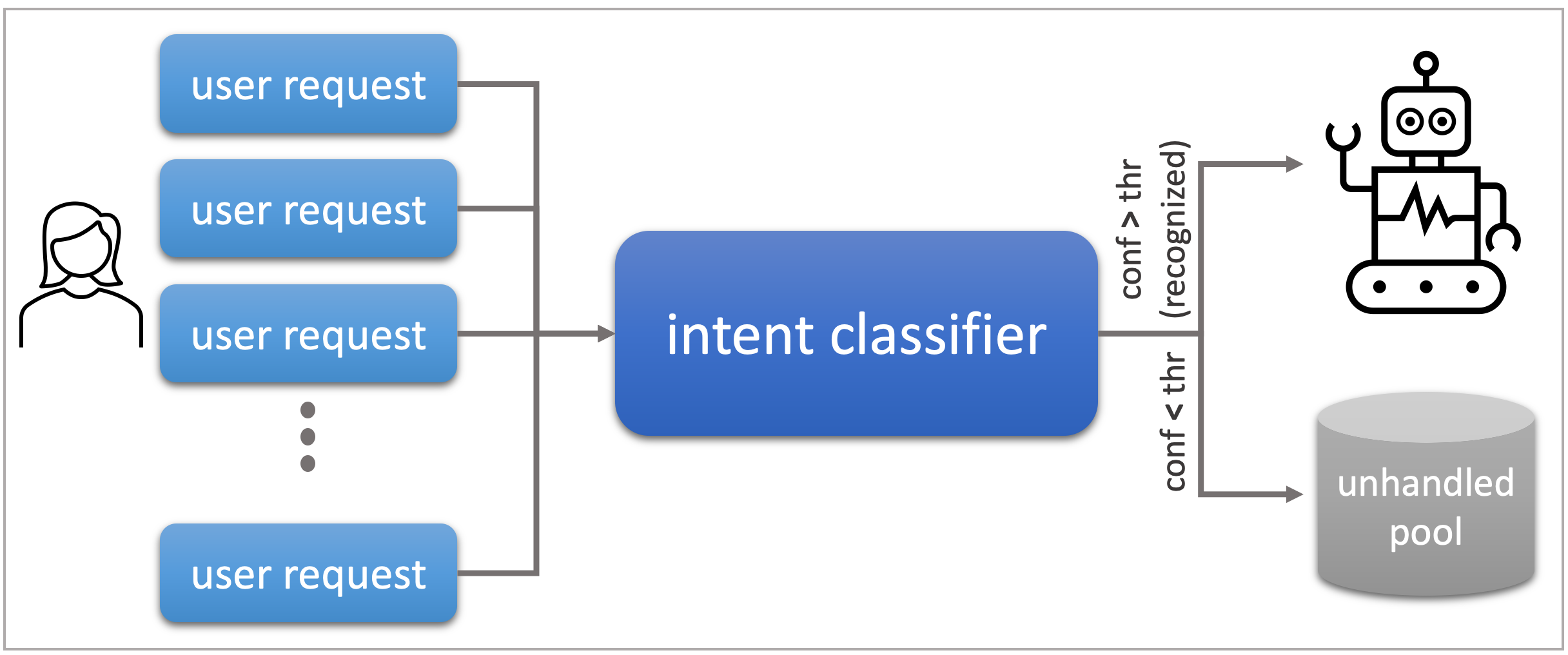}
}
\caption{Natural language understanding (NLU) module. Based to the intent classifier's confidence level, first user utterances are `recognized' and associated with an execution flow, or stored in an unhandled pool.}
\label{fig:nlu-module}
\end{figure}

Goal-oriented dialog systems often fail to recognize the intent of natural language requests due to system errors, incomplete service coverage, or insufficient training \citep{grudin2019chatbots, kvale2019improving}.\footnote{In most virtual assistants, a user utterance is considered unhandled by the system's NLU module either by design (often referred to as ``out-of-domain''), or due to the system's failure to attribute the utterance to one of its existing intents.} In practice, these cases are normally identified using intent classifier uncertainty. Here, user utterances that are predicted to have a level of confidence below a certain threshold to any of the predefined intents, are identified and reported as unrecognized or \textit{unhandled}. Figure~\ref{fig:nlu-module} presents the NLU module from a typical task-oriented dialog system: the user utterance is either transformed into an intent with an appropriate flow of subsequent actions, or labeled as unrecognized and stored in the \textit{unhandled pool} (Figure~\ref{fig:nlu-module}).

Unhandled utterances often carry over various aspects of potential importance, including novel examples of existing intents, novel topics that may introduce a new intent, or seasonal topical peaks that should be monitored but not necessarily modeled within the system. In large deployments, the number of unhandled utterances can reach tens of thousands on a daily basis. Despite their evident importance for continuous bot improvement, tools for gaining effective insights into unhandled utterances have not been developed sufficiently, leaving a vast body of knowledge, as well as a range of potentially actionable items, unexploited.

Gaining insights into the topical distribution of unrecognized requests can be achieved using unsupervised text analysis tools, such as clustering or topic modeling. Indeed, identifying clusters of semantically similar utterances can surface topics of interest to a conversation analyst. We show that traditional clustering algorithms result in sub-optimal performance due to the unique traits of unhandled utterances in dialog systems: an unknown number of expected clusters and a very long tail of outliers. Consequently, we propose and evaluate a radius-based variant of the k-means clustering algorithm \cite{lloyd1982least}, that does not require a fixed number of clusters and tolerates outliers gracefully. We show that it outperforms its out-of-the-box counterparts on a range of real-world customer, as well as public datasets. The algorithm has recently been evaluated on the task of intent discovery in the context of large-scale, \href{https://vaxchat.org/}{production chatbot}, being ranked first (out of $4$) at coverage metrics, and second at utterance partitioning \cite{gretz2022benchmark}.

We propose an end-to-end pipeline for surfacing topical clusters in unhandled user requests, including utterance cleanup, a designated clustering procedure and its extensive evaluation, a novel approach to cluster representatives extraction, and cluster naming. We approach this task in a real-world setting of commercial task-oriented dialog systems, and demonstrate the benefits of the suggested approach on multiple publicly available, as well as proprietary, datasets. %
%
\section{Clustering of Unrecognized Requests}
\label{sec:clustering}

Consider a virtual assistant aimed to attend to public questions about Covid-19. The rapidly evolving situation with the pandemic means that novel requests are likely to be introduced to the bot on a daily basis. As such, changes in international travel regulations would entail requests related to PCR test availability, and the decision to offer booster shots for seniors might cause a spike in questions about vaccine appointments for elderly citizens. Monitoring and prompt detection of these topics are fundamental for continuous bot improvement.

\subsection{Clustering Utterances}
Here we describe the main clustering procedure followed by an optional single merging step.

\subsubsection{Main Clustering Procedure}
\begin{figure}
\centering
\resizebox{\columnwidth}{!}{
\includegraphics{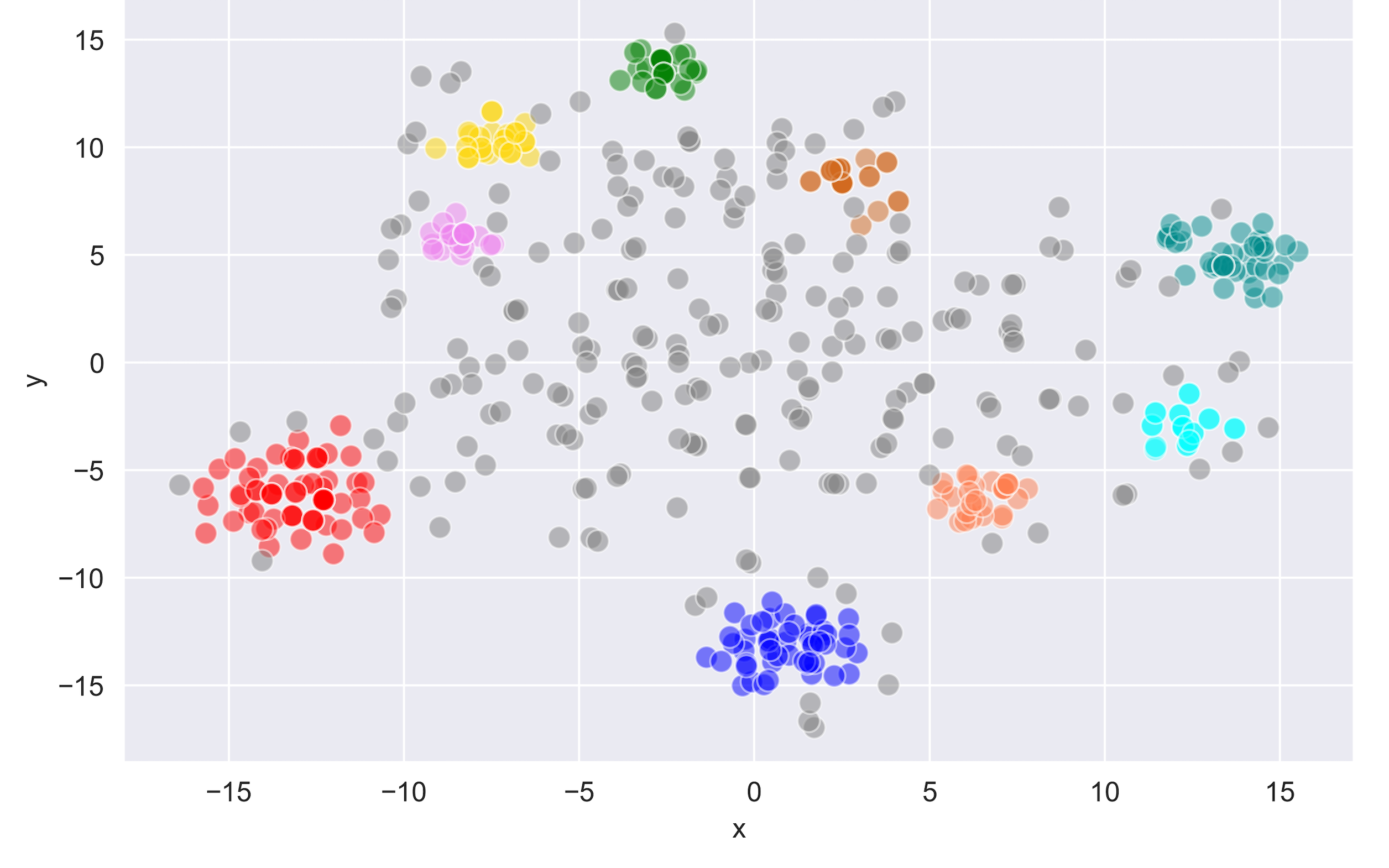}
}
\caption{t-SNE projection of a sample of unrecognized user requests in a production task-oriented dialog system. Identified clusters are in color, outliers -- in grey.}
\label{fig:clustering-space}
\end{figure}

\paragraph{Clustering requirements} Multiple traits make up an effective clustering procedure in our scenario. First, the number of clusters is unknown, and has to be discovered by the clustering algorithm. Second, the nature of data typically implies several large and coherent clusters, where users repeatedly introduce very similar requests, and a very long tail of unique (often noisy) utterances that do not have similar counterparts. While the latter are of somewhat limited importance, they can amount to a significant ratio of the input data. There is an evident trade-off between the size of the generated clusters, their density or sparsity, and the number of outliers: smaller and denser clusters entail larger amounts of outliers. The decision regarding the precise outcome granularity may vary according to domain and bot maturity. Growing deployments, with a high volume of unrecognized requests, could benefit from surfacing large and coarse topics that are subject to automation. That said, mature deployments are likely to focus on fine-grained coherent clusters of utterances, introducing enhancements into the existing solution. Our third requirement is, therefore, a configurable density of the outcome clusters, which can be set up prior to the clustering procedure.
Figure~\ref{fig:clustering-space} illustrates a typical outcome of the clustering process; identified clusters are depicted in color, while the outliers, which constitute approximately half of the instances, appear in grey.

Existing clustering solutions can be roughly categorized across two major dimensions in terms of functional requirements: those requiring a fixed number of output clusters (1.a) and those that do not (1.b); those forcing cluster assignment on the entire dataset (2.a) and those tolerating outliers (2.b). Our clustering solution should accommodate  (1.b) and (2.b): the number of clusters is determined by the clustering procedure, allowing for outliers. DBSCAN \citep{ester1996density} and its descendant variants constitute a popular family of clustering solutions that satisfies these requirements; we, therefore, evaluate our algorithm against implementations of DBSCAN and its hierarchical version HDBSCAN \citep{mcinnes2017hdbscan}.

\paragraph{Data representation} Given a set of $m$ unhandled utterances ${U}${=}($u_1$,$u_2$, ...,$u_m$), we compute their vector representations ${E}${=}($e_1$,$e_2$, ...,$e_m$) using a sentence encoder. Multiple available encoders were evaluated for this purpose, considering both effectiveness and efficiency (see Section~\ref{sec:eval-approach}).

\paragraph{Radius-based clustering (RBC)} We introduce a variant of the popular k-means clustering algorithm, complying with our clustering requirements by (1) imposing a strict cluster assignment criterion and (2) eventually omitting points that do not constitute clusters exceeding a predefined size. 

Specifically, we iterate over randomly-ordered vectors in $E$, where each utterance vector can be assigned to an existing cluster if certain conditions are satisfied; otherwise, it initiates a new cluster. In order to join an existing cluster, the utterance is required to surpass a predefined similarity threshold $min\_sim$ with the cluster's centroid,\footnote{Following the k-means notation, we compute a cluster's centroid as the arithmetic mean of its member vectors.} implying its placement within a certain \textit{radius} from the centroid. If multiple clusters satisfy the similarity requirement, the utterance is assigned to the cluster with the highest proximity i.e., the cluster with the highest semantic similarity to its centroid. Additional iterations over the utterances are further performed, re-assigning them to different clusters if needed, until convergence, or until a pre-defined number of iterations is exhausted.\footnote{Contrary to k-means, our algorithm is not sensitive to its (random) initialization, since we are not required to select K centroids; utterance processing order has shown only negligible effect on the final outcome.} The amount of clusters generated by the final partition is controlled by the $min\_size$ value: elements that constitute clusters of small size (in particular, those with a single item) are considered outliers. Algorithm~\ref{alg:clustering} presents the algorithm's pseudo-code.

\SetKwInOut{Input}{input}
\SetKwInOut{Output}{output}
\SetKwComment{Comment}{/*}{*/}

\begin{algorithm}
\caption{Radius-based Clustering}
\label{alg:clustering}
\SetInd{0.75em}{0.25em}
\small

\textbf{input}: {E (e1, e2, ... en)} \Comment{\hspace{0.05cm} elements}
\textbf{input}: {min\_sim} \Comment{\hspace{0.05cm} min similarity}
\textbf{input}: {min\_size} \Comment{\hspace{0.05cm} min cluster size}
\BlankLine

$C \gets \emptyset$

\While{convergence criteria are not met}{
    \BlankLine
    \For{each element $e_i{\in}E$} {
        \BlankLine
        \eIf{the highest similarity of $e_i$ to any existing cluster exceeds min\_sim}{
            \BlankLine
            assign $e_i$ to its most similar cluster $c$ \\
            re-calculate the centroid of $c$
            \BlankLine
        }{
            \BlankLine
            create a new cluster $c^\prime$ and assign $e_i$ to it \\
            set the centroid of $c^\prime$ to be $e_i$ \\
            add $c^\prime$ to $C$
            \BlankLine
            
        }
    }
}
\BlankLine
\Comment{clusters with fewer elements than the predefined $min\_size$ are considered outliers}
\BlankLine
\textbf{return}: each $c{\in}C$ of size exceeding $min\_size$
\end{algorithm}

\subsubsection{Cluster Merging}
\label{sec:cluster-merging}

Cluster merging has been extensively used as a means to determine the optimal clustering outcome in the scenario where the `true' number of partitions is unknown \citep{krishnapuram1994generation, kaymak2002fuzzy, xiong2004similarity}. These start with a large number of clusters and iteratively merge compatible partitions until the optimization criteria is satisfied. Beginning with fine-grained partitioning, we perform an (optional) single step of cluster merging, combining similar clusters into larger groups. A similar outcome could potentially be obtained by relaxing the \textit{min\_sim} similarity threshold and thereby, generating more heterogeneous flat clusters in the first place. However, a single step of cluster merging yielded results that outperform flat clustering on a range of datasets (see Table~\ref{tbl:evaluation-results} and Section~\ref{sec:evaluation-results} for details). 
Classical agglomerative hierarchical clustering (AHC) algorithms merge pairs of lower-level clusters by minimizing the agglomerative criterion: a similarity requirement that has to be satisfied for a pair of clusters to be merged. Similar to AHC, we seek to merge clusters exhibiting high mutual similarity. In contrast to AHC, our approach is not pair-wise, rather it constitutes a subsequent invocation of the RBC that takes embeddings of the flat cluster centroids as its input. 

Formally, given a set of clusters $C$ of size $k{=}|C|$, identified by the algorithm, we compute the set of cluster centroid vectors $E^c${=}($e^c_1$,$e^c_2$, ...,$e^c_k$); these vectors are assumed to reliably represent the semantics of their corresponding clusters. 
$E^c$ is further used as an input to subsequent invocation of the RBC algorithm, where the \textit{min\_sim} parameter can possibly differ from the previous invocation.

\begin{table*}[hbt]
\centering
\resizebox{\textwidth}{!}{
\begin{tabular}{l|l}
cluster name: \textbf{difference covid flu} (28) & cluster name: \textbf{covid pregnancy} (17) \\ \hline
is covid the same as the flu? (4) & covid 19 and pregnancy (10) \\ 
how is covid different from the flu? (3) & covid risks for a pregnant woman (4) \\
what is the difference between covid 19 and flu? & what is the risk of covid for pregnant women? \\
what's the difference between covid and flu & is covid-19 dangerous when pregnant? \\
is the covid the same as cold? & 7 months pregnant and tested positive for covid, any risks?  \\
covid vs flu vs sars & covid 19 during pregnancy \\
\end{tabular}
}
\caption{Example clusters of user requests generated by the RBC algorithm when applied on the Covid-19 dataset. Only a partial list of cluster members is presented in the table; the number in parenthesis denotes a cluster size.}
\label{tbl:clustering-example}
\end{table*}

\paragraph{Example Clustering Result} Table~\ref{tbl:clustering-example} presents two example clusters generated from user requests to the Covid-19 bot. We applied the main RBC clustering procedure and a single subsequent merge step. Semantically related utterances are grouped together, where the number beside an utterance reflects its frequency in the cluster. As a concrete example, `is covid the same as the flu?' was asked four times by different users.

\subsection{Evaluation}
\label{sec:clustering-evaluation}

We performed a comparative evaluation of the proposed clustering algorithm and HDBSCAN\footnote{DBSCAN resulted in outcomes systematically inferior to HDBSCAN; hence, it was excluded from further experiments.}, using common clustering evaluation metrics. The nature of the topical distribution of unrecognized utterances is probably most closely resembled by dialog systems \textit{intent classification} datasets, where semantically similar training examples are grouped into classes, based on their intent. We used these classes to simulate cluster partitioning for the purpose of evaluation.
We make use of three publicly available intent classification datasets (\citet{liu2019benchmarking}, \citet{larson2019evaluation} and \citet{tepper2020balancing}), as well as three datasets from real-world task-oriented chatbots in the domains of telecom, finance and retail. Table \ref{tbl:datasets} presents the datasets details.

\begin{table}[hbt]
\centering
\resizebox{\columnwidth}{!}{
\begin{tabular}{l|rrrr}
dataset & intents  & examples & mean & STD \\ \hline
\citet{liu2019benchmarking} & 46 & 20849 & 453.23 & 896.34 \\
\citet{larson2019evaluation} & 150 & 22500  & 150.00 & 0.00 \\
\citet{tepper2020balancing} & 57 & 844  & 14.80 & 14.16 \\ \hline
telecom    & 167   & 6364  & 38.10 & 26.74 \\
finance    & 142   & 2301  & 16.20 & 25.28 \\
retail    & 103   & 1714  & 16.64 & 11.42 \\
\end{tabular}
}
\caption{Datasets details: the number of intents, total training examples, mean and STD of the num of examples. We excluded out-of-scope examples from the \citet{larson2019evaluation} dataset for the sake of evaluation.}
\label{tbl:datasets}
\end{table}

\subsubsection{Evaluation Approach}
\label{sec:eval-approach}
The main approaches to clustering evaluation include extrinsic methods, which assume a ground truth, and intrinsic methods, which work in the absence of ground truth. Extrinsic techniques compare the clustering outcome to a human-generated \textit{gold standard} partitioning. Intrinsic techniques assess the resulting clusters by measuring characteristics such as cohesion, separation, distortion, and likelihood \citep{pfitzner2009characterization}. We employ two popular extrinsic and intrinsic evaluation metrics: adjusted random index (ARI, \citep{hubert1985comparing}) and Silhouette Score \citep{rousseeuw1987silhouettes}. We vary the parameters of the RBC algorithm: merge type with none vs. single step (see Section~\ref{sec:cluster-merging}); the encoder used for distance matrix construction: the SentenceTransformer (ST) encoder \citep{reimers2019sentence} vs. the Universal Sentence Encoder (USE) \citep{cer2018universal}; min similarity threshold used as a cluster ``radius'' was optimized on a held-out set of intents, per dataset. Both ARI and Silhouette yield values in the [-1, 1] range, where -1, 0 and 1 mean incorrect, arbitrary, and perfect assignment, respectively.
The unique nature of our clustering requirements introduces a challenge to standard extrinsic evaluation techniques. Specifically, the min cluster size attribute controls the number of outliers, by considering only clusters that exceed the minimum number of members (see Figure~\ref{fig:clustering-space}). 
Aiming to mimic the ground truth partition (i.e, the intent classification datasets), we set the \textit{min\_size} attribute to the minimal class size in the dataset, subject to evaluation. As such, this attribute was set to $150$ for the \citet{larson2019evaluation} dataset, but to $2$ for the finance dataset.

Both evaluation techniques assume full partitioning of the input space. Therefore, for our evaluation, we exclude the set outliers generated by our clustering algorithm altogether: only the subset of instances constructing the outcome clusters (e.g., instances depicted in color in Figure~\ref{fig:clustering-space}) was used to compute both ARI and Silhouette. For completeness, we also report the ratio of a dataset utterances covered by the generated partition (`\% clst' in Table~\ref{tbl:evaluation-results}), where the higher, the better.

\subsubsection{Evaluation Results}
\label{sec:evaluation-results}

Table~\ref{tbl:evaluation-results} presents the results of our evaluation.
Clearly, the RBC algorithm outperforms HDBSCAN across the board for both ARI and Silhouette scores, with the exception of the retail dataset, where the second best ARI score ($0.37$) is obtained by RBC along with over $80$\% of clustered utterances (compared to only $49.79$\% by HDBSCAN). HDBSCAN also outperforms RBC in terms of the ratio of clustered utterances for~\citet{liu2019benchmarking} and the telecom dataset. However, these results are achieved by a nearly arbitrary partition of the input data, as mirrored by the extremely low ARI and Silhouette scores. We conclude that RBC outperforms its out-of-the-box counterpart on virtually all datasets in this work.
The ratio of clustered examples (`\% clst') exhibits considerable variance among the datasets; this result is indicative of the varying levels of semantic coherence of the underlying intent classes, which are typically constructed manually by a bot designer. As such, over $87$\% of all training examples were covered by the clustering procedure for the retail dataset, but only $33.90$\% for \citet{larson2019evaluation}. 


The extremely poor results obtained for the telecom dataset by HDBSCAN stem from its clustering outcome that only contains two clusters: (1) a small group of unique examples and (2) all the rest.

\begin{table}[hbt]
\centering
\resizebox{\columnwidth}{!}{
\begin{tabular}{l|l|rrrr|rr}
& algo & \multicolumn{4}{c|}{RBC} & \multicolumn{2}{c}{HDBSCAN} \\ \hline
& merge type & \multicolumn{2}{c}{no merge} & \multicolumn{2}{c|}{single step} & \multicolumn{2}{c}{---} \\ \hline
& encoder & \multicolumn{1}{c}{USE} & \multicolumn{1}{c}{ST} & \multicolumn{1}{c}{USE} & \multicolumn{1}{c|}{ST} & \multicolumn{1}{c}{USE} & \multicolumn{1}{c}{ST} \\ \hline
\multirow{3}{*}{\STAB{\rotatebox[origin=c]{90}{{Liu}}}}
& ARI           & 0.42  & 0.40  & \textbf{0.74}  & 0.44  & 0.42  & 0.03  \\
& Silhouette    & 0.47  & 0.42  & \textbf{0.67}  & 0.50  & 0.39  & 0.09  \\
& \% clst  & 12.12 & 12.03 & 12.12 & 16.09 & 12.69 & \textbf{38.36} \\ \hline

\multirow{3}{*}{\STAB{\rotatebox[origin=c]{90}{{Larson}}}}
& ARI           &  \textbf{0.89}   & 0.86  & 0.68  & 0.76  & 0.49  & 0.69  \\
& Silhouette    & 0.47 & \textbf{0.50} & 0.48  & \textbf{0.50}  & 0.39  & 0.47  \\
& \% clst  & 16.29 & 32.60 & 16.29 & \textbf{33.90} & 24.92 & 32.98 \\ \hline

\multirow{3}{*}{\STAB{\rotatebox[origin=c]{90}{{Tepper}}}}
& ARI           &  0.66   & 0.65  & \textbf{0.73}  & 0.52  & 0.69  & 0.67  \\
& Silhouette    & 0.45   & 0.49  & \textbf{0.51}  & 0.37  & 0.45  & 0.46  \\
& \% clst  & 79.68  & 85.12 & 79.68 & \textbf{88.18} & 58.31 & 60.15 \\ \hline \hline

\multirow{3}{*}{\STAB{\rotatebox[origin=c]{90}{{telecom}}}}
& ARI           & 0.32 & 0.54 & \textbf{0.63} & 0.38 & 0.00 & 0.00 \\
& Silhouette    & 0.17 & \textbf{0.20} & 0.18 & 0.11 & 0.00 & 0.00 \\
& \% clst  & 25.18 & 46.87 & 25.18 & 59.78 & 83.24 & \textbf{97.90} \\ \hline

\multirow{3}{*}{\STAB{\rotatebox[origin=c]{90}{{finance}}}}
& ARI           & 0.40 & 0.42 & 0.45 & \textbf{0.56} & 0.45 & 0.49 \\
& Silhouette    & 0.37 & \textbf{0.39} & 0.35 & 0.32 & 0.34 & 0.35 \\
& \% clst  & 47.59 & 63.26 & 47.59 & \textbf{74.28} & 23.46 & 36.22 \\ \hline

\multirow{3}{*}{\STAB{\rotatebox[origin=c]{90}{{retail}}}}
& ARI           & 0.28 & 0.37 & 0.31 & 0.24 & 0.37 & \textbf{0.38} \\
& Silhouette    & 0.24 & 0.28 & 0.22 & 0.27 & 0.23 & \textbf{0.32} \\
& \% clst  & 68.19 & 80.43 & 68.19 & \textbf{87.18} & 37.55 & 49.79 \\

\end{tabular}
}
\caption{Clustering evaluation results; `\% clst' denotes the ratio of clustered examples out of total; the best result in a row is boldfaced.}
\label{tbl:evaluation-results}
\end{table}

\paragraph{Runtime and Memory}
Due to its nearly polynomial complexity, the proposed clustering algorithm may entail efficiency considerations for a very large amount of data. As such, with pre-computed request embeddings, clustering $20$K unhandled requests results in less than 10 seconds, while clustering $85$K requests takes $82$ seconds with over $850$MB of RAM consumption. All experiments were conducted on a server with $8$ CPUs.

\section{Selecting Cluster Representatives}
\label{sec:representatives}

Contemporary large-scale deployments of virtual assistants must cope with increasingly high volumes of incoming user requests. A typical large task-oriented system can accept over $100$K requests (i.e., user utterances) per day, where the amount of conversations that pass the initial step of intent identification varies between 40\% and 80\%. Consequently, tens of thousands of requests can be identified as unrecognized on a daily basis. Clustering these utterances would result in large clusters that are often impractical for manual processing. Providing conversation analysts with a limited set of \textit{cluster representatives} is a fundamental step toward extracting value from the unrecognized data.

\subsection{Representative Characteristics}
A plausible set of representative cluster utterances has to satisfy two desirable properties: utterance \textit{centrality} and \textit{diversity}. We define an utterance centrality to be proportional to its frequency in a cluster: requests with higher frequency should be boosted, since they are typical of the way people express their needs to the bot. The diversity of the utterance set mirrors the subtle differences in the phrasing and meaning of utterances; these reflect the various ways people can express the same need.

Sampling randomly from a cluster may result in a sub-optimal set of representatives, in terms of both centrality and diversity. Consider the example where no `covid 19 and pregnancy' requests (Table \ref{tbl:clustering-example}, right) are selected as representatives (low centrality), or both `what is the difference between covid 19 and flu?' and `what's the difference between covid and flu' (Table \ref{tbl:clustering-example}, left) are selected (low diversity). Contrary to these examples, the set \{`is covid the same as the flu?', `is the covid the same as cold?', `covid vs flue vs sars'\} contains utterance of high centrality (the first utterance), and comprehensive coverage of the entire cluster semantics.

\subsection{Selecting Representatives}
To ensure diversity and centrality among the selected representatives, we use determinantal point process (DPP). Specifically, we consider a restricted class of DPPs known as L-ensembles. Given a set of items, $\mathcal{S}$, L-ensembles define a probability distribution over the power set of $\mathcal{S}$. Equivalently, L-ensembles define a probability distribution over binary vectors of length $|\mathcal{S}|$, where the $i^{th}$ entry in the vector indicates if the $i^{th}$ item in $\mathcal{S}$ was included in the subset or not. These indicator variables are negatively correlated where the correlations are governed by a positive semidefinite matrix $K$. L-ensembles ensure that the more similar two items are, as indicated by the corresponding entry in the kernel matrix, the less likely are they to occur in the same sampled subset. Thus, it is an excellent model for ensuring diversity among the selected representatives. 

Given a positive semi-definite kernel matrix $K$, the probability of $A{\subset}\mathcal{S}$ is governed in an L-ensemble as $P(A) \propto det(K_A)$,
%
%
where $K_A$ is the restriction of $K$ to the indices present in the subset $A$. We construct the kernel matrix to ensure that samples from the L-ensemble have high centrality while also being diverse. To achieve this, we first project the embeddings of the utterances within the cluster onto a unit sphere. We further take into consideration the factor of centrality by scaling the vectors' length based on their frequency in the cluster. Given the resultant embeddings $E$, where the embedding of the $i^{th}$ entry is the $i^{th}$ row vector in $E$, the kernel matrix is obtained by $K{=}EE^{\mathrm{T}}$. Thus, the $(i,j)^{th}$ entry of the kernel corresponds to the angle between the $i^{th}$ and $j^{th}$ vector scaled by the frequency of occurrence of those vectors. We make use of the freely available \href{https://github.com/guilgautier/DPPy}{DPPy Python package} for sampling a subset of representatives, given the above kernel matrix.

\paragraph{Evaluation}
Using the clustering approach in Section~\ref{sec:clustering} we extracted $50$ clusters of varying sizes of unhandled user requests from a large-scale production system. A set of three cluster representatives was extracted using the technique described in this section, along with two baselines: (1) three random cluster members, (2) three unique most frequent cluster members. Three in-house annotators labeled their preferred alternative, satisfying \textit{centrality} and \textit{diversity} properties in the best way. The majority vote was obtained in $47$ out of $50$ cases, with $37$ out of $47$ ($79$\%) choices preferring the centrality-diversity approach. The mean pairwise Cohen's Kappa between the annotators was $0.44$.
\section{Cluster Naming}
\label{sec:naming}

Assigning cluster with names, or labels, is an essential step toward their consumability. Common approaches to this task resort to simple but reliable techniques based on word n-gram extraction, such as \textit{tf-idf}; many of these techniques made their way into the first large-scale information retrieval (IR) systems \citep{ramos2003using, aizawa2003information}. Here, we distinguish between the task of cluster naming (extracting a coherent phrase reliably reflecting a cluster's content) and the task of keyword extraction (providing a sequence of one or more words for a compact representation of a document).

Common approaches to cluster naming include extracting one of the cluster's members to reflect the cluster's content; extracting such a member can be done by by naively selecting the most frequent member in the cluster or by choosing a member satisfying maximum cosine similarity to the cluster's centroid \citep{alicante2016semantic}. In other cases, a good name may not occur directly as one of the cluster's members, and hence requires different handling. Some works were trying to investigate the contribution of external knowledge-bases for cluster naming, by incorporating Wikipedia pages' meta-data corresponding to the cluster's content \citep{carmel2009enhancing}, while others were trying to generate clusters' queries, as a mixture of cluster-internal and differential labeling \citep{hagen2015query}.
Contemporary large pretrained large language models can also be used for the task of keyword extraction. Here we make use of KeyBERT -- an approach based on BERT \citep{devlin2018bert} -- for identifying key phrases in a cluster, and evaluate the outcome against tf-idf.

\paragraph{Cluster Labeling with tf-idf} 
We treat all utterances in individual clusters from a set $C{=}(c_1, c_2, ..., c_k)$ as distinct documents. We first applied lemmatization to these documents using the \href{https://spacy.io/}{spacy toolkit} \citep{spacy2}, excluded stopwords, and further ranked all ngram token sequences of length $N$ (for $N{\in}(1,2,3)$) by their tf-idf score. 
The ngram with the highest score was selected as the cluster name.

\paragraph{Cluster Labeling with KeyBERT} 
Treating each cluster as a document, we first extract document-level representation using a pretrained BERT language model.\footnote{We use `all-MiniLM-L6-v2' model in our experiments.} 
We further extract ngram representations for all unique word ngrams in the document, and compute semantic similarity between each ngram's embedding and that of the document. Ngram with the highest cosine similarity to the document is selected as the cluster name.\footnote{We make use of the freely available \href{https://github.com/MaartenGr/KeyBERT}{KeyBERT} package.}

\vspace{-0.025in}
\paragraph{Evaluation}
Adhering to the same evaluation paradigm as Section~\ref{sec:clustering-evaluation}, we use the six intent classification datasets for assessing the quality of cluster naming techniques. A common practice for building an intent training dataset involves assigning each class in the training set with a meaningful name, typically mirroring the semantics of the class. As such, an intent class grouping example requests about Covid-19 testing information in \citet{tepper2020balancing}, is named `testing information'. For each class in the intent training set, we compare the automatically extracted class name to that assigned to the class by the dataset creator, where the similarity is obtained by encoding the two phrases -- the original class name and the candidate one -- and computing their cosine similarity.

Table~\ref{tbl:cluster-naming} presents the results for the two methods. Neither approach systematically outperforms the other, and the only significant difference in favor of the tf-idf approach is found for \citet{liu2019benchmarking}. We, therefore, conclude that the two approaches are roughly comparable and adhere to the faster tf-idf method in our pipeline solution.

\begin{table}[hbt]
\centering
\begin{tabular}{l|lc}
dataset & tf-idf & KeyBERT \\ \hline
\citet{liu2019benchmarking} & \hspace{0.3em}\textbf{0.718}* & 0.626 \\
\citet{larson2019evaluation} & \hspace{0.3em}\textbf{0.555} & 0.489 \\
\citet{tepper2020balancing} & \hspace{0.3em}\textbf{0.481} & 0.460 \\ \hline
telecom   & \hspace{0.3em}0.437 & \textbf{0.470}  \\
finance   & \hspace{0.3em}\textbf{0.438} & 0.426  \\
retail    & \hspace{0.3em}0.375 & \textbf{0.393}  \\
\end{tabular}
\caption{Cluster naming evaluation: for each dataset, the mean pairwise similarity between the predefined intent name and the assigned keyphrase is presented. `*' denotes significant difference at p-val{\textless}0.01 using the Wilcoxon (Mann–Whitney) ranksums test. }
\label{tbl:cluster-naming}
\end{table}

\vspace{-0.08in}
\section{Conclusions and Future Work}
\label{sec:conclusions}

Analyzing unrecognized user requests is a fundamental step toward improving task-oriented dialog systems. We present an end-to-end pipeline for clustering, representatives selection, and cluster naming -- procedures that facilitate the effective and efficient exploration of utterances unrecognized by the NLU module. We propose a clustering variant of the popular k-means algorithm, and show that outperforms its out-of-the-box alternatives on multiple metrics. We also suggest a novel approach to extracting representative utterances while simultaneously optimizing their centrality and diversity.

Our future work includes the evaluation of our clustering approach with additional datasets, exploration of additional approaches to representative set selection, and advanced techniques for cluster naming. Leveraging clustering results to automatically identify actionable recommendations for conversation analyst is another venue of significant practical importance, we plan to pursue.

\section{Ethical Considerations}
\label{sec:ethical}

Cluster representative sets (Section~\ref{sec:representatives}) were annotated by in-house workers who were compensated with above minimum wages. To protect user privacy, no personally identifiable information (e.g., name, address) were presented to the annotators.

\section*{Acknowledgements}
We are grateful to the anonymous reviewers and the meta reviewer for their constructive feedback. We would also like to thank Chani Sacharen for her kind help with earlier versions of this work.

\bibliographystyle{acl_natbib}
\bibliography{main}

\begin{thebibliography}{26}
\expandafter\ifx\csname natexlab\endcsname\relax\def\natexlab#1{#1}\fi

\bibitem[{Aizawa(2003)}]{aizawa2003information}
Akiko Aizawa. 2003.
\newblock \href
  {https://www.sciencedirect.com/science/article/abs/pii/S0306457302000213} {An
  {I}nformation-{T}heoretic {P}erspective of tf--idf {M}easures}.
\newblock \emph{Information Processing \& Management}, 39(1):45--65.

\bibitem[{Alicante et~al.(2016)Alicante, Corazza, Isgr{\`o}, and
  Silvestri}]{alicante2016semantic}
Anita Alicante, Anna Corazza, Francesco Isgr{\`o}, and Stefano Silvestri. 2016.
\newblock \href
  {https://link.springer.com/chapter/10.1007/978-3-319-39687-3_18} {Semantic
  cluster labeling for medical relations}.
\newblock In \emph{International Conference on Innovation in Medicine and
  Healthcare}, pages 183--193. Springer.

\bibitem[{Carmel et~al.(2009)Carmel, Roitman, and
  Zwerdling}]{carmel2009enhancing}
David Carmel, Haggai Roitman, and Naama Zwerdling. 2009.
\newblock \href
  {https://dl.acm.org/doi/abs/10.1145/1571941.1571967?casa_token=z1bASeKTp8YAAAAA:UKzLRehN9J8WRAV_ikHsTs_KqrFDCrzchj3uxw-ZOMZOmRioym7v7IHEAVcsXDd1b37e0ir7UMKt}
  {Enhancing cluster labeling using wikipedia}.
\newblock In \emph{Proceedings of the 32nd international ACM SIGIR conference
  on Research and development in information retrieval}, pages 139--146.

\bibitem[{Cer et~al.(2018)Cer, Yang, Kong, Hua, Limtiaco, John, Constant,
  Guajardo-C{\'e}spedes, Yuan, Tar et~al.}]{cer2018universal}
Daniel Cer, Yinfei Yang, Sheng-yi Kong, Nan Hua, Nicole Limtiaco, Rhomni~St
  John, Noah Constant, Mario Guajardo-C{\'e}spedes, Steve Yuan, Chris Tar,
  et~al. 2018.
\newblock \href {https://arxiv.org/abs/1803.11175} {{U}niversal {S}entence
  {E}ncoder}.
\newblock \emph{arXiv preprint arXiv:1803.11175}.

\bibitem[{Chen et~al.(2017)Chen, Liu, Yin, and Tang}]{chen2017survey}
Hongshen Chen, Xiaorui Liu, Dawei Yin, and Jiliang Tang. 2017.
\newblock \href {https://dl.acm.org/doi/abs/10.1145/3166054.3166058} {A
  {S}urvey on {D}ialogue {S}ystems: {R}ecent {A}dvances and {N}ew {F}rontiers}.
\newblock \emph{Acm Sigkdd Explorations Newsletter}, 19(2):25--35.

\bibitem[{Devlin et~al.(2018)Devlin, Chang, Lee, and
  Toutanova}]{devlin2018bert}
Jacob Devlin, Ming-Wei Chang, Kenton Lee, and Kristina Toutanova. 2018.
\newblock \href {https://arxiv.org/abs/1810.04805} {Bert: Pre-training of deep
  bidirectional transformers for language understanding}.
\newblock \emph{arXiv preprint arXiv:1810.04805}.

\bibitem[{Ester et~al.(1996)Ester, Kriegel, Sander, and Xu}]{ester1996density}
Martin Ester, Hans-Peter Kriegel, J{\"o}rg Sander, and Xiaowei Xu. 1996.
\newblock \href {https://www.aaai.org/Papers/KDD/1996/KDD96-037.pdf}
  {{D}ensity-{B}ased {S}patial {C}lustering of {A}pplications with {N}oise}.
\newblock In \emph{Int. Conf. Knowledge Discovery and Data Mining}, volume 240,
  page~6.

\bibitem[{Gretz et~al.(2022)Gretz, Toledo, Friedman, Lahav, Weeks, Bar-Zeev,
  Sedoc, Sangha, Katz, and Slonim}]{gretz2022benchmark}
Shai Gretz, Assaf Toledo, Roni Friedman, Dan Lahav, Rose Weeks, Naor Bar-Zeev,
  Jo{\~a}o Sedoc, Pooja Sangha, Yoav Katz, and Noam Slonim. 2022.
\newblock \href {https://arxiv.org/abs/2205.11966} {Benchmark data and
  evaluation framework for intent discovery around covid-19 vaccine hesitancy}.
\newblock \emph{arXiv preprint arXiv:2205.11966}.

\bibitem[{Grudin and Jacques(2019)}]{grudin2019chatbots}
Jonathan Grudin and Richard Jacques. 2019.
\newblock \href {https://dl.acm.org/doi/abs/10.1145/3290605.3300439}
  {{C}hatbots, {H}umbots, and the {Q}uest for {A}rtificial {G}eneral
  {I}ntelligence}.
\newblock In \emph{Proceedings of the 2019 CHI Conference on Human Factors in
  Computing Systems}.

\bibitem[{Hagen et~al.(2015)Hagen, Michel, and Stein}]{hagen2015query}
Matthias Hagen, Maximilian Michel, and Benno Stein. 2015.
\newblock \href
  {https://link.springer.com/chapter/10.1007/978-3-319-19581-0_10} {What was
  the query? generating queries for document sets with applications in cluster
  labeling}.
\newblock In \emph{International Conference on Applications of Natural Language
  to Information Systems}, pages 124--133. Springer.

\bibitem[{Honnibal and Montani(2017)}]{spacy2}
Matthew Honnibal and Ines Montani. 2017.
\newblock \href {https://sentometrics-research.com/publication/72/} {{spaCy 2}:
  {N}atural {L}anguage {U}nderstanding with {B}loom {E}mbeddings,
  {C}onvolutional {N}eural {N}etworks and {I}ncremental {P}arsing}.
\newblock \emph{Sentometrics Research}.

\bibitem[{Huang et~al.(2020)Huang, Zhu, and Gao}]{huang2020challenges}
Minlie Huang, Xiaoyan Zhu, and Jianfeng Gao. 2020.
\newblock \href {https://dl.acm.org/doi/abs/10.1145/3383123} {{C}hallenges in
  {B}uilding {I}ntelligent {O}pen-{D}omain {D}ialog {S}ystems}.
\newblock \emph{ACM Transactions on Information Systems (TOIS)}, 38(3):1--32.

\bibitem[{Hubert and Arabie(1985)}]{hubert1985comparing}
Lawrence Hubert and Phipps Arabie. 1985.
\newblock \href {https://link.springer.com/article/10.1007/BF01908075}
  {{C}omparing {P}artitions}.
\newblock \emph{Journal of classification}, 2(1).

\bibitem[{Kaymak and Setnes(2002)}]{kaymak2002fuzzy}
Uzay Kaymak and Magne Setnes. 2002.
\newblock \href {https://ieeexplore.ieee.org/abstract/document/1097771}
  {{F}uzzy {C}lustering with {V}olume {P}rototypes and {A}daptive {C}luster
  {M}erging}.
\newblock \emph{IEEE Transactions on Fuzzy Systems}, 10(6):705--712.

\bibitem[{Krishnapuram(1994)}]{krishnapuram1994generation}
Raghu Krishnapuram. 1994.
\newblock \href {https://ieeexplore.ieee.org/abstract/document/343851}
  {{G}eneration of {M}embership {F}unctions via {P}ossibilistic {C}lustering}.
\newblock In \emph{Proceedings of 1994 IEEE 3rd International Fuzzy Systems
  Conference}, pages 902--908. IEEE.

\bibitem[{Kvale et~al.(2019)Kvale, Sell, Hodnebrog, and
  F{\o}lstad}]{kvale2019improving}
Knut Kvale, Olav~Alexander Sell, Stig Hodnebrog, and Asbj{\o}rn F{\o}lstad.
  2019.
\newblock \href
  {https://link.springer.com/chapter/10.1007/978-3-030-39540-7_13} {{I}mproving
  {C}onversations: {L}essons {L}earnt from {M}anual {A}nalysis of {C}hatbot
  {D}ialogues}.
\newblock In \emph{International workshop on chatbot research and design},
  pages 187--200. Springer.

\bibitem[{Larson et~al.(2019)Larson, Mahendran, Peper, Clarke, Lee, Hill,
  Kummerfeld, Leach, Laurenzano, Tang et~al.}]{larson2019evaluation}
Stefan Larson, Anish Mahendran, Joseph~J Peper, Christopher Clarke, Andrew Lee,
  Parker Hill, Jonathan~K Kummerfeld, Kevin Leach, Michael~A Laurenzano,
  Lingjia Tang, et~al. 2019.
\newblock \href {https://aclanthology.org/D19-1131/} {An {E}valuation dataset
  for intent classification and out-of-scope prediction}.
\newblock In \emph{Proceedings of the 2019 Conference on Empirical Methods in
  Natural Language Processing and the 9th International Joint Conference on
  Natural Language Processing (EMNLP-IJCNLP)}, pages 1311--1316.

\bibitem[{Liu et~al.(2019)Liu, Eshghi, Swietojanski, and
  Rieser}]{liu2019benchmarking}
Xingkun Liu, Arash Eshghi, Pawel Swietojanski, and Verena Rieser. 2019.
\newblock \href
  {https://researchportal.hw.ac.uk/en/publications/benchmarking-natural-language-understanding-services-for-building}
  {{B}enchmarking {N}atural {L}anguage {U}nderstanding {S}ervices for
  {B}uilding {C}onversational {A}gents}.
\newblock In \emph{10th International Workshop on Spoken Dialogue Systems
  Technology 2019}, volume 714, pages 165--183. Springer.

\bibitem[{Lloyd(1982)}]{lloyd1982least}
Stuart Lloyd. 1982.
\newblock \href {https://ieeexplore.ieee.org/abstract/document/1056489}
  {{L}east {S}quares {Q}uantization in {PCM}}.
\newblock \emph{IEEE transactions on information theory}, 28(2):129--137.

\bibitem[{McInnes et~al.(2017)McInnes, Healy, and Astels}]{mcinnes2017hdbscan}
Leland McInnes, John Healy, and Steve Astels. 2017.
\newblock \href {https://joss.theoj.org/papers/10.21105/joss.00205} {hdbscan:
  {H}ierarchical {D}ensity {B}ased {C}lustering}.
\newblock \emph{Journal of Open Source Software}, 2(11):205.

\bibitem[{Pfitzner et~al.(2009)Pfitzner, Leibbrandt, and
  Powers}]{pfitzner2009characterization}
Darius Pfitzner, Richard Leibbrandt, and David Powers. 2009.
\newblock \href {https://link.springer.com/article/10.1007/s10115-008-0150-6}
  {{C}haracterization and {E}valuation of {S}imilarity {M}easures for {P}airs
  of {C}lusterings}.
\newblock \emph{Knowledge and Information Systems}, 19(3):361--394.

\bibitem[{Ramos et~al.(2003)}]{ramos2003using}
Juan Ramos et~al. 2003.
\newblock \href
  {https://citeseerx.ist.psu.edu/viewdoc/download?doi=10.1.1.121.1424&rep=rep1&type=pdf}
  {{U}sing tf-idf to {D}etermine {W}ord {R}elevance in {D}ocument {Q}ueries}.
\newblock In \emph{Proceedings of the first instructional conference on machine
  learning}, volume 242, pages 29--48. Citeseer.

\bibitem[{Reimers and Gurevych(2019)}]{reimers2019sentence}
Nils Reimers and Iryna Gurevych. 2019.
\newblock \href {https://aclanthology.org/D19-1410/} {{S}entence-{BERT}:
  {S}entence {E}mbeddings using {S}iamese {BERT}-{N}etworks}.
\newblock In \emph{Proceedings of the 2019 Conference on Empirical Methods in
  Natural Language Processing and the 9th International Joint Conference on
  Natural Language Processing (EMNLP-IJCNLP)}, pages 3982--3992.

\bibitem[{Rousseeuw(1987)}]{rousseeuw1987silhouettes}
Peter~J Rousseeuw. 1987.
\newblock \href
  {https://www.sciencedirect.com/science/article/pii/0377042787901257}
  {{S}ilhouettes: a {G}raphical {A}id to the {I}nterpretation and {V}alidation
  of {C}luster {A}nalysis}.
\newblock \emph{Journal of computational and applied mathematics}, 20:53--65.

\bibitem[{Tepper et~al.(2020)Tepper, Goldbraich, Zwerdling, Kour, Tavor, and
  Carmeli}]{tepper2020balancing}
Naama Tepper, Esther Goldbraich, Naama Zwerdling, George Kour, Ateret~Anaby
  Tavor, and Boaz Carmeli. 2020.
\newblock \href {https://aclanthology.org/2020.findings-emnlp.130/}
  {{B}alancing via {G}eneration for {M}ulti-{C}lass {T}ext {C}lassification
  {I}mprovement}.
\newblock In \emph{Findings of the Association for Computational Linguistics:
  EMNLP 2020}, pages 1440--1452.

\bibitem[{Xiong et~al.(2004)Xiong, Chan, and Tan}]{xiong2004similarity}
Xuejian Xiong, Kap~Luk Chan, and Kian~Lee Tan. 2004.
\newblock \href {https://dl.acm.org/doi/abs/10.5555/1036843.1036917}
  {{S}imilarity-{D}riven {C}luster {M}erging {M}ethod for {U}nsupervised
  {F}uzzy {C}lustering}.
\newblock In \emph{Proceedings of the 20th conference on Uncertainty in
  artificial intelligence}, pages 611--618.

\end{thebibliography}

\end{document}